\documentclass[sn-mathphys]{sn-jnl}

\jyear{2022}%

\theoremstyle{thmstyleone}%
\theoremstyle{thmstyletwo}%

\theoremstyle{thmstylethree}%

\begin{document}

\title[Single-stage Rotate Object Detector via Two Points with Solar Corona Heatmap]{Single-stage Rotate Object Detector via Two Points with Solar Corona Heatmap}


\author[1]{ \sur{Beihang Song}}\email{beihangsongwhu@163.com}
\author*[1]{ \sur{Jing Li}}\email{leejingcn@whu.edu.cn}

\author[2]{\sur{Shan Xue}}\email{sxue@uow.edu.au}
\author[1]{\sur{Jun Chang}}\email{chang.jun@whu.edu.cn}
\author[3]{\sur{Jia Wu}}\email{jia.wu@mq.edu.au}
\author[1]{\sur{Jun Wan}}\email{junwan2014@whu.edu.cn}
\author[1]{\sur{Tianpeng Liu}}\email{tianpengliu@whu.edu.cn}



\affil[1]{\orgdiv{School of Computer Science}, \orgname{Wuhan University}, \orgaddress{ \city{wuhuan}, \postcode{430072}, \country{China}}}

\affil[2]{\orgdiv{School of Computing and Information Technology}, \orgname{University of Wollongong}, \orgaddress{\country{Australia}}}

\affil[3]{\orgdiv{School of Computing}, \orgname{Macquarie University},  \city{Sydney}, \state{NSW2109}, \country{Australia}}

\abstract{Oriented object detection is a crucial task in computer vision. Current top-down oriented detection methods usually directly detect entire objects, and not only neglecting the authentic direction of targets, but also do not fully utilise the key semantic information, which causes a decrease in detection accuracy. In this study, we developed a single-stage rotating object detector via two points with a solar corona heatmap (ROTP) to detect oriented objects. The ROTP predicts parts of the object and then aggregates them to form a whole image. Herein, we meticulously represent an object in a random direction using the vertex, centre point with width, and height. Specifically, we regress two heatmaps that characterise the relative location of each object, which enhances the accuracy of locating objects and avoids deviations caused by angle predictions. To rectify the central misjudgement of the Gaussian heatmap on high-aspect ratio targets, we designed a solar corona heatmap generation method to improve the perception difference between the central and non-central samples. Additionally, we predicted the vertex relative to the direction of the centre point to connect two key points that belong to the same goal. Experiments on the HRSC 2016, UCASAOD, and DOTA datasets show that our ROTP achieves the most advanced performance with a simpler modelling and less manual intervention.}

\keywords{Oriented Object , Heatmap , Key Points , Object Detection}



\maketitle

\section{Introduction}\label{sec1}

\begin{figure}[H]
    \includegraphics[scale=0.6]{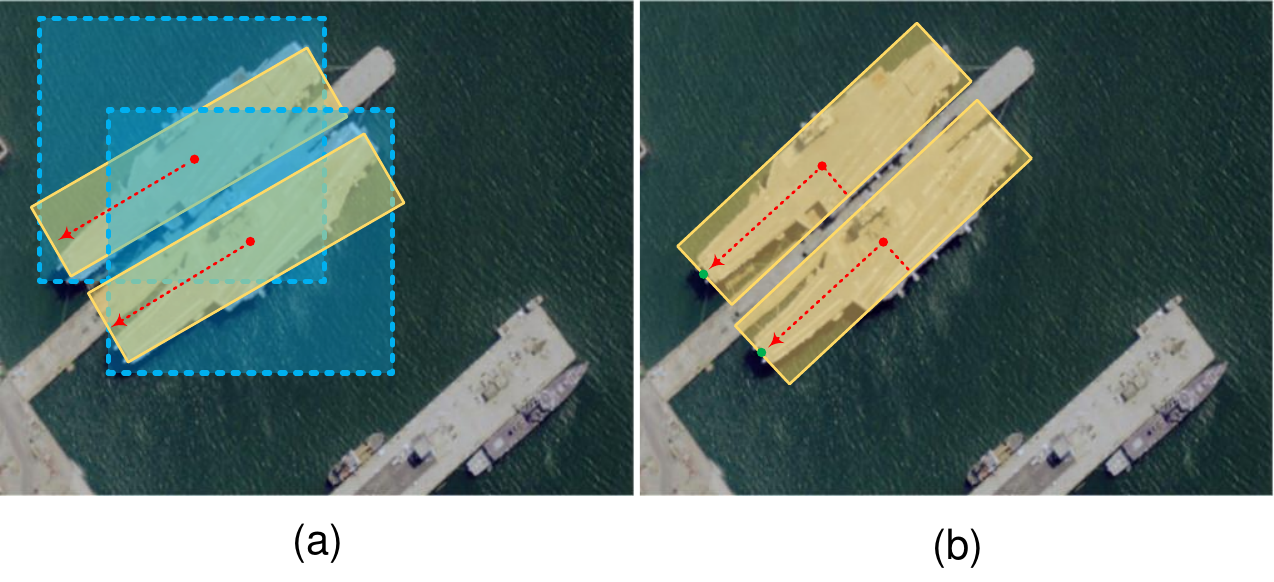}
    \centering
    \caption{(a) Deviation that inevitably occurs when only the angle is predicted in rotating target detection. (b) Addition of predicted top position, which enhances accuracy of bounding boxes.
}
\label{fig1}
\end{figure}

Object detection is a basic and challenging task in computer vision. CNN algorithms are required to obtain each instance of interest in an image, as well as bounding boxes with category labels. Currently, object detection algorithms have shifted from conventional methods to deep learning-based algorithms, such as YOLO\cite{redmon2016you}, SSD\cite{liu2016ssd}, and Fast RCNN\cite{girshick2015fast}. These algorithms have achieved significant success in the detection of horizontal objects. However, if an object inclines significantly, as shown in Figure \ref{fig1} (a), the bounding box detected by these horizontal detectors contains a lot of background, which decreases the accuracy when describing the object information. Moreover, when multiple objects of the same category are very close, the horizontal bounding box will include parts of multiple objects, which may be misjudged as the same object in the later processing stage of the model, leading to misdetection. To overcome these problems, existing studies\cite{jiang2017r2cnn,ma2018arbitrary,liao2017textboxes,liao2018textboxes++,zhou2020arbitrary,2021Oriented} have modified the mainstream horizontal object detector to detect oriented objects, including by increasing predictions of angle or position of four corner points directly to represent oriented objects. Although better results have been achieved, some issues remain. These issues can be caused either by the horizontal object detector or by defects in the model design. The detection performance of an anchor-based detector\cite{ma2018arbitrary,liao2017textboxes,xu2020gliding} has certain requirements regarding the size, aspect ratio, and quantity of anchors. Further, although the ratio and aspect ratio of the anchors remain unchanged after initialisation, even with elaborate design, it is difficult to deal with candidates with large shape changes, specifically for targets in oriented objects. Moreover, a predefined positioning framework can hinder the generalisation ability of the detector. For example, the YOLOv3\cite{redmon2018yolov3}  anchor clustering based on different data cannot be applied to objects with large sizes or reduced aspect ratios. To achieve a higher average precision rate, the anchor-based detector RRPN was established\cite{ma2018arbitrary}. Due to differences in angles, anchors must be placed densely on the input image, providing many parameters for model calculation. However, these anchors increase the number of negative samples, which leads to an imbalance between the positive and negative samples for training. Although in other methods\cite{zhou2020arbitrary,2021Oriented,zhou2017east} the directly predicted angle reached a high mAP, they still have the problem of inaccurate angle prediction. More specifically, the angle difference between the predicted and real values should be minimal. Although the slant border based on this angle can locate an object, as shown in Figure \ref{fig1} (a), the accuracy of the final model is reduced.

In this study, we developed a single-stage rotating object detector via two points with a solar corona heatmap (ROTP). Traditional detectors use an anchor-based mechanism based on the regression angle, object size, and corner coordinates. In contrast to the two modes of most popular oriented detectors, our proposed ROTP adopts two key points to detect targets. As shown in Figure \ref{fig1} (b), by locating the two key points of the vertex and the centre, as well as predicting the size of the target, a simple mapping function can be used to generate a slanted rectangular border of the oriented objects. To achieve this, the ROTP outputs five feature maps, including a heatmap of the centre point and vertex, and predicts the target centre and vertex positions using keypoint detection. The other feature maps indicate the long and short sides of a target, the offsets of the key points, and the direction of the vertex relative to the central point.

Our contributions in this paper can be summarized as follows:

\begin{enumerate}
    \item We propose a single-stage rotating object detector via two points with a solar corona heatmap (ROTP). ROTP uses the predicted vertex and centre points to describe the oriented object, which avoids the regression error caused by direct prediction of the angle. It also replaces angle prediction with prediction of the relative direction, thus improving the accuracy to describe the orientation of objects.
    \item We designed a solar corona heatmap (SCH) based on the spatial position relationship to predict the centre point more accurately for slender objects. This method considers the large aspect ratio of remote sensing objects, which has a more robust performance than Gaussian heatmaps. 
    \item Our method obtains preferable precision with more semantic design and fewer hyperparametric designs compared with other classical-oriented object detection methods based on deep learning.
\end{enumerate}

The rest of this paper is organised as follows: Section 2 reviews related work in the field of horizontal object detection and oriented object detection. Section 3 presents the overall framework of the proposed ROTP model in detail, as well as the function design. Implementation details, comparison experiments, and ablation studies are presented in Section 4, then conclusions are drawn in Section 5. 

\section{Related Work}\label{sec2}

Many excellent oriented object detectors are based on horizontal object detectors. In this section, we first introduce horizontal object detection methods and then introduce oriented object detection methods.

\noindent\textbf{Horizontal object detection.}
Object detection aims to detect each object in natural scene images with a rectangular border. Deep learning, which has a broad range of applications \cite{DBLPXing,9565320,IJCAI20Fanzhen}, has already been successfully applied in object detection. Current classical deep learning-based methods can be summarised as single-stage and two-stage methods. The single-stage method directly predicts all objects, whereas the two-stage method is subdivided based on preliminary predictions. Therefore, although the two-stage method is more accurate, it is also slower. These methods can be further categorised into anchor-based and anchor-free methods. Specifically, the anchor-based two-stage model represented by RCNN\cite{2014}, Fast RCNN\cite{girshick2015fast} and faster RCNN\cite{2016Faster} set multiscale anchor boxes in advance, which can be understood as a series of candidate regions of scales and sizes. First, they generate ROIs and then use features of the ROI to predict object categories. SSD\cite{liu2016ssd}, YOLOv2\cite{redmon2017yolo9000} and its variants\cite{redmon2018yolov3} are representative single-stage anchor-based methods. There is no candidate ROI area, and the classification and regression results are regressed in the final stage. Specifically, SSD\cite{liu2016ssd} combines the advantages of two-stage and single-junction segments to achieve a balance between speed and precision. Subsequently, Retinanet\cite{lin2017focal} and DSSD\cite{2017DSSD} are used to fuse multiscale features in the model to further improve the detection accuracy. In addition, anchor-free methods are becoming popular, including FoveaBox\cite{2019FoveaBox}, RepPoints\cite{2019RepPoints} and FCOS\cite{tian2019fcos}based on per-pixel detection, while heatmaps were introduced in \cite{law2018cornernet} and Centernet\cite{zhou2019objects} for object detection to achieve object-to-key mapping. These detection methods simplify the network structure by removing the anchors and improving the detection speed, and provide a new research direction to detect objects. Nevertheless, the abovementioned object detection methods only generate position information in the horizontal or vertical directions, which limits their universality. For example, in scene text and other rotating object images, the aspect ratio of the instance is relatively large, the arrangement is dense, and the direction is arbitrary, which requires more accurate position information to describe these objects. Therefore, oriented object detection has gradually become a popular research direction.

\noindent\textbf{Oriented object detection.}
Owing to the huge scale changes and arbitrary directions, detecting oriented targets is challenging. Extensive research has been devoted to this task. Many oriented object detectors have been proposed based on horizontal object detectors. RRPN\cite{ma2018arbitrary} and $ R^2CNN $\cite{jiang2017r2cnn} are the classical methods. Based on Fast RCNN\cite{girshick2015fast}, $ R^2CNN $ \cite{jiang2017r2cnn} raises two pooling sizes and an output branch to predict the corner positions. RRPN\cite{ma2018arbitrary} achieved better prediction results by adding rotating anchors at different angles. Based on
SSD\cite{liu2016ssd}, textboxes\cite{liao2017textboxes} and textboxes++\cite{liao2018textboxes++} have been proposed. According to the characteristics of slender text lines, a rectangular convolution kernel was proposed, and the results of the instance are regressed by the vortices. RRD\cite{2018Rotation} predicts a rotating object based on the invariance property of rotation features, improving the accuracy of regression of long text. EAST\cite{zhou2017east} had a U-shaped network\cite{2015U} and predicted the border with angle and the four corner positions of the instance simultaneously. The above methods are applicable to the field of scene text detection, whereas aerial remote sensing target detection is more difficult. Compared with text lines, remote sensing targets have many categories, such as complex background, multiscale, and a large number of dense small targets. Many robust oriented target detectors have emerged based on horizontal detectors. For example, ROI-Transformer\cite{ding2018learning} extracts rotation-invariant features on the ROI, which increases the accuracy of the next step of classification and regression. ICN\cite{azimi2018towards} combines pyramid and feature pyramid modules and achieves good results on remote sensing datasets. The glide vertex\cite{xu2020gliding} predicts the deviation of the four angled corner points at the horizontal border to obtain a more accurate direction bounding box. P-RSDet\cite{zhou2020arbitrary} and BBVector\cite{2021Oriented} introduced the anchor free heatmap detection method into oriented target detection to achieve a fast and accurate detection effect.

The above anchor free rotation detector still uses the method of horizontal target mapping heatmap, and does not take into account the problem that Gaussian heatmap is prone to position perception deviation on high aspect ratio targets.

\begin{figure}[H]
    \includegraphics[scale=0.6]{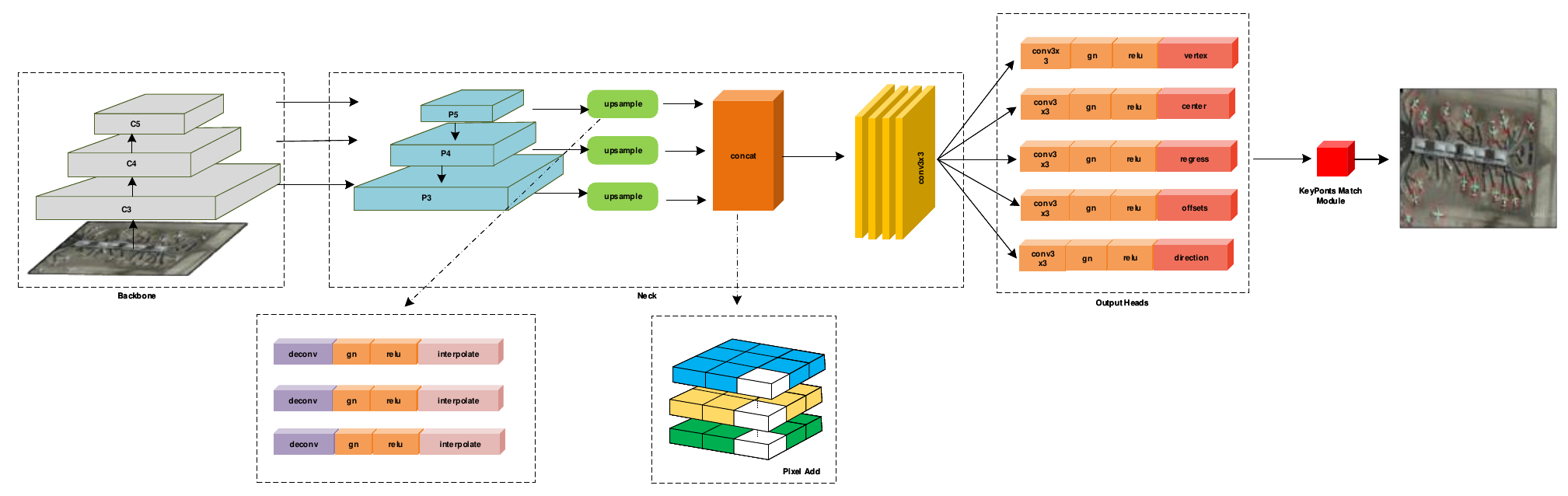}
    \centering
    \caption{ROTP architecture. The ROTP is divided into three parts: backbone, neck, and head. When an image with the size of WxHx3 is input into our model, it will output five feature maps, among which two are heatmaps for predicting the centre points and vertices, and the other three are regression for the targets’ size, coordinates bias, and direction, respectively. Finally, all key points are matched to obtain the final output. 
}
\label{fig2}
\end{figure}

\section{Proposed Method}\label{sec3}
Herein, we first elaborate on the instance representation of our method and then describe the SCH generation. Finally, we present details of the proposed ROTP and its loss function. 

\begin{figure}[H]
    \includegraphics[scale=0.6]{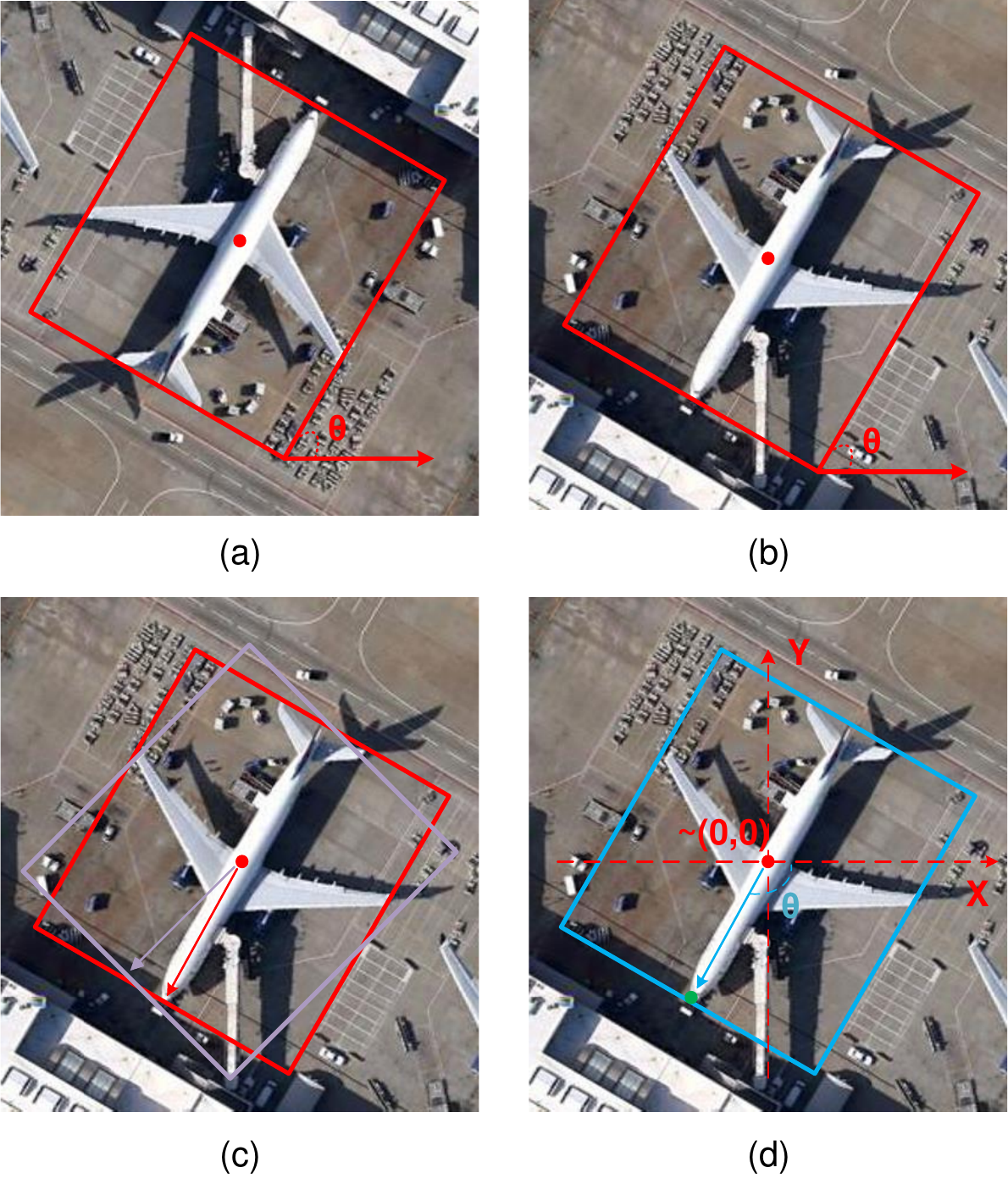}
    \centering
    \caption{Current work only considers the angle of the boundary box relative to the image coordinate system, as shown in (a) and (b), but does not consider the real orientation of the object. The red border in (c) represents the groundtruth, and the purple border is the predicted. The predicted border covers the whole target, but the IOU between it and the groundtruth seems large. As shown in (d), a coordinate system was established for each target. The centre point of the target is coordinate point zero, and the angle represents the included angle of the vector between the vertex and the centre point relative to the positive direction of the X-axis.
}
\label{fig3}
\end{figure}
\subsection{Instance Representation}\label{subsec1}
Oriented objects can generally be expressed by two patterns, which can be used interchangeably. The first type of pattern is represented by  $ (x_1, y_1) $ , $ (x_2, y_2) $ , $ (x_3, y_3) $ , and $ (x_4, y_4) $ in the order of four clockwise corners, and the second type is $ (x, y, w, h, \alpha) $, where $ (x, y) $ represents the center point, w and h are the width and height of the target, $ \alpha $ represents the rotation angle of the target.In this study, oriented objects are modelled with two key points: $ (\sum_{i=1}^{4}{x_i/4,}\sum_{i=1}^{4}{y_i/4}) $ , which is the centre point $ (x_c, y_c) $ , and $ (\sum_{i=1}^{2}{x_i/2,}\sum_{i=1}^{2}{y_i/2}) $ , which is the vertex $ (x_t, y_t) $. 

From the introduction in the previous section, we use the vertex, centre point, and length and width to represent a target. First, $ (x_c^i, y_c^i) $ , which represents the coordinate of the centre point from the heatmap, is extracted. Here, $ i $ represents the index of the targets. Thereafter,  $ (x_t^i, y_t^i) $, which represents the positions of all vertices, is extracted. Because datasets are labelled with the problem of inaccuracy, the vertex label is occasionally marked on the background area, and the training input for the vertex coordinates is 0.9 times the coordinates along the centre point. To combine the centre point and the corresponding vertices, we added a prediction head in order to predict the direction of the vertices relative to the centre point, as shown in Figure \ref{fig3} (a) and (b). Previous studies only focused on the target bounding box angle in the image coordinate system and but ignored the targets; although this approach decreases the predicted angle range, the value of angle loss value may suddenly increase during the calculation, \cite{2019Learning}. In this paper, we propose a relative direction that is not based on an image coordinate system but is rather based on our well-designed coordinate system of the centre of the object. As shown in Figure \ref{fig3} (d), the relative direction can be expressed as the included angle between the vector from the vertex to the centre and with respect to the positive X-axis. The relative direction is calculated using the following formula:
\begin{small}
    \begin{equation}
        \theta_i=\left\{\begin{matrix}
            \frac{180}{\pi}\arccos\frac{{x_t}^i-{x_c}^i}{\sqrt{({x_t}^i-{x_c}^i)^2 + ({y_t}^i-{y_c}^i})^2}&{x_t}^i-{x_c}^i\geq 0,{y_t}^i-{y_c}^i\geq 0  \\
            \frac{180}{\pi}\arccos\frac{{x_t}^i-{x_c}^i}{\sqrt{({x_t}^i-{x_c}^i)^2 + ({y_t}^i-{y_c}^i})^2}&{x_t}^i-{x_c}^i > 0,{y_t}^i-{y_c}^i > 0  \\
            360 - \frac{180}{\pi}\arccos\frac{{x_t}^i-{x_c}^i}{\sqrt{({x_t}^i-{x_c}^i)^2 + ({y_t}^i-{y_c}^i})^2}&{x_t}^i-{x_c}^i\leq 0,{y_t}^i-{y_c}^i\leq 0  \\
            360 - \frac{180}{\pi}\arccos\frac{{x_t}^i-{x_c}^i}{\sqrt{({x_t}^i-{x_c}^i)^2 + ({y_t}^i-{y_c}^i})^2}&{x_t}^i-{x_c}^i > 0,{y_t}^i-{y_c}^i < 0  \\
          \end{matrix}\right.
    \end{equation}
    \end{small}

The relative direction $ \theta_i $ is defined as ranging from 0 to 360. We first establish the vector from the centre point $ (x_c,y_c) $ to the vertex $ (x_t,y_t) $, and then calculate the cosine of the positive angle $ \alpha $ between the vector and the X-axis according to Eq\ref{eqcosa}. We then obtain the radian through the inverse trig function . Finally, according to the position of the vertex relative to the central point, the vertex is mapped to the four quadrants with the central point as the origin, and the relative direction is determined. The advantage of this method is that the angle is regarded as a constant with the same scale as that of the target, and the regression prediction can be directly performed without additional processing.
\begin{equation}
        \cos\alpha=\frac{{x_t}-{x_c}}{\sqrt{({x_t}-{x_c})^2 + ({y_t}-{y_c}})^2} 
        \label{eqcosa}
\end{equation}
In the training stage, the direction of the tagets was optimized with a smooth $ L_1 $ loss as follows:
\begin{equation}
    L_{D} = \frac{1}{N}\sum_{i=1}^NSmooth_{L_1}(\theta_i - \acute{\theta_i})
\end{equation}
Where $ N $ is the number of whole peak elements, $ \theta_i $ refers to the direction of instance, $ \acute{\theta_i} $ denotes the prediction of direction, and i denotes the index of all objects in a batch. The smooth $ L_1 $ loss is represented as follows:
\begin{equation}
    f(x) = \left\{\begin{matrix}
        0.5x^2&|x|<1 \\
        |x| - 0.5 & otherwise\\
    \end{matrix}\right.
\end{equation}

\begin{figure}[H]
    \includegraphics[scale=0.6]{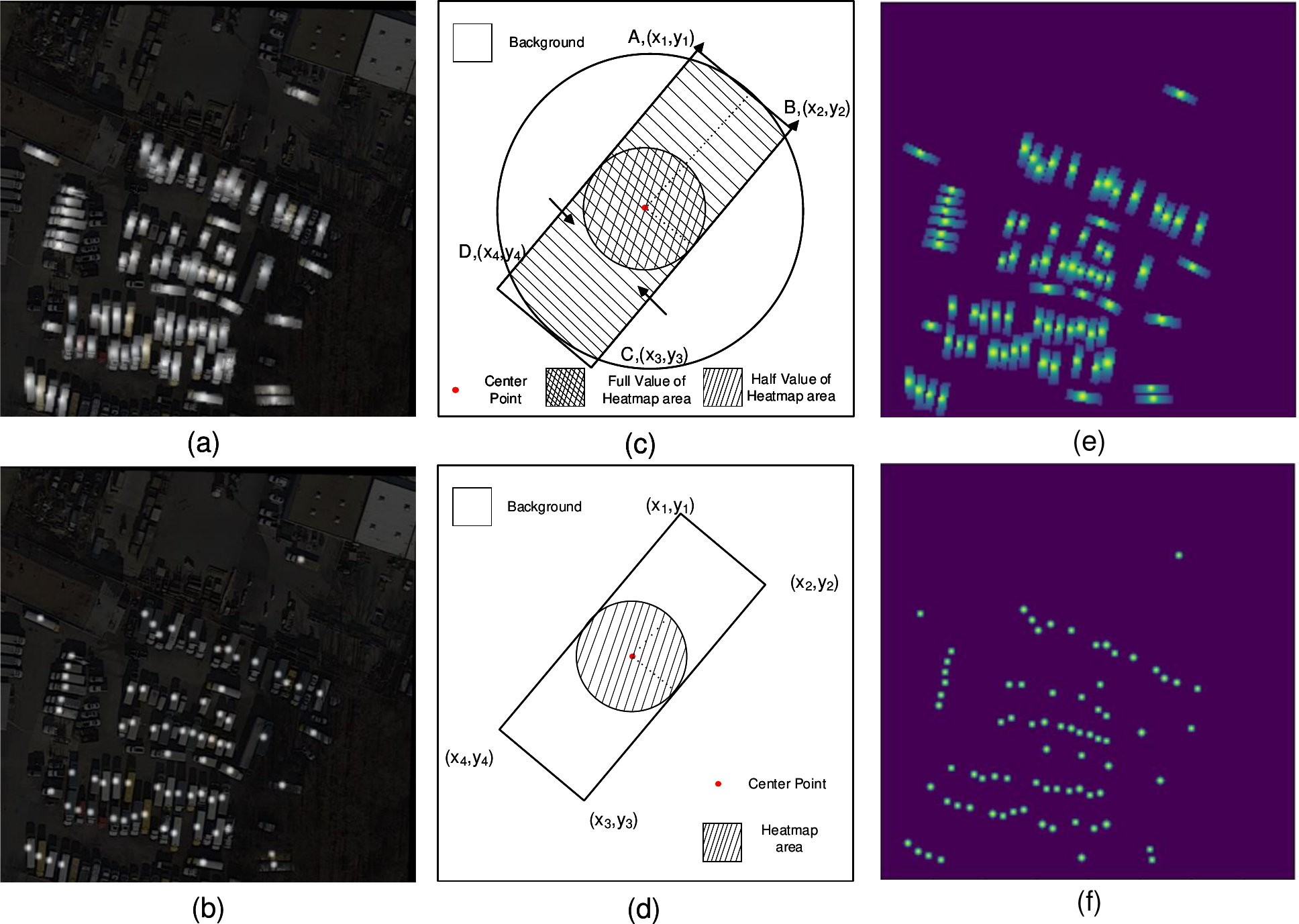}
    \centering
    \caption{Large quantities of edge feature information are lost in the Gaussian heatmap when describing a rotating target with a large aspect ratio, as shown in (b) and (d). When the object is very narrow, the model can only learn a small amount of information available near the centre point. Our heatmap method, as shown in (a) and (c), enables the model to learn information at the edges of objects while learning information around the centre point. Panels (e) and (f) show the heatmap with colours.}
    \label{fig4}
\end{figure}

\subsection{Solar Corona Heatmap}\label{subsec2}
Existing studies on key point detections have mostly used the Gaussian kernel function, $ Y_{hm}=\exp^{-\frac{(x-x_p)^2+(y-y_p)^2}{2\delta^2}} $ to mapping objects into heatmap $ Y\in {[0,1]}^{\frac{W}{d}\times\frac{H}{d}\times C}  $, where $ (x_c, y_c) $ represents the coordinates of the centre point, $ (x_p,y_p) $ is an element on heatmap, and $ \sigma $ is the radius of the Gaussian kernel. When mapping a remote-sensing object into a Gaussian heatmap, as shown in Figure \ref{fig4} (b) and (d), the heatmap area only accounts for a small part of the ground truth bounding box when the target is extremely slender (for example, large trucks, bridges, and some designated ships). In the case of a large aspect ratio, the radius of the heatmap is small, and the resulting area would be small.  For long and block objects without obvious texture changes, the detector may miscalculate the peak value at the top or tail during reasoning, which is detrimental to oriented object detection.  Such defects are also reflected in the horizontal heatmap detector; thus, the detector should be able to better learn the peak value of the heatmap. Centernet\cite{zhou2019objects} performs many data enhancement operations, such as cutting, reversing, and translating, to ensure the accuracy of the detector. For remote sensing images, as the objects are small and dense, such data enhancement operations will lower the accuracy of the detector. Therefore, this study proposes a spatial perception heatmap to address this problem, as shown in Figure \ref{fig4} (a). As shown in Figure \ref{fig4} (c), the heatmap radius based on half of the longer side can include all objects. In the scenario of dense objects, this method judges the peak value of the surrounding objects into a low confidence level area. Thus, our solar corona heatmap is based on both the short and long sides; therefore, it retains the weights of the surrounding area of the centre point and perceives the area of the head and tail. The formula used is as follows:

\begin{equation}
    H_c = \frac{1}{2}(e^{-\frac{(x_p-x_c)^2+(y_p-y_c)^2}{\mu_1 \times h}}+e^{-\frac{(x_p-x_c)^2+(y_p-y_c)^2}{\mu_1 \times w}})\ (x_p,y_p)\in T 
\end{equation}
\begin{equation}
    H_t = e^{-\frac{(x_p-x_t)^2+(y_p-y_t)^2}{\mu_1 \times h}}
\end{equation}

Where $ H_c $ is the heatmap of center point, $ H_t $ means the heatmap of vertex point, $ w $ is the side length with the smallest difference from the value of twice the distance from the vertex to the centre point, and $ h $ is another side length. $ T $ represents the entire point-group of an instance. In this study, we used $ \mu = 0.125 $ in all the experiments. In the training stage, only the peak points were positive. All other points, including those in the Gaussian bulge, were negative. The model learns only one point, and the loss function does not usually converge because the number of positive samples is not sufficient compared with the number of background samples. To deal with this issue, Eq\ref{eqhml} is used. The closer the target is to the peak point, the higher the confidence is, and the points beyond a certain range are, the lower the confidence is ; thus, the gap between the peak point and other points can be measured effectively. The variant focal loss is used to calculate the ground truth and prediction.

\begin{equation}
    L_{hm}=-\frac{1}{N_{pos}}\sum_{ps}\left\{\begin{matrix}
        (1-\rho_{ps})^\alpha\log(\rho_{ps})&\acute{\rho_{ps}} = 1 \\
        (1-\acute{\rho_{ps}})^\beta(\rho_{ps})^\alpha\log(1 - \rho_{ps})&\acute{\rho_{ps}} \neq 1 \\
      \end{matrix}\right.
      \label{eqhml}
\end{equation}
where $ N_{pos} $ represents the number of pixels covered by the heatmap, $ \rho_{ps} $ represents the predicted confidence, $ \acute{\rho_{ps}} $ represents the ground truth, $ \alpha $ and $ \beta $ are adjustable parameters that are used to ensure that the model can better learn the pixels with different confidence levels; their values are set to 2 and 4, respectively, in this study.
\subsection{Offset}\label{subsec3}
When in the inference stage, the position of the maximum  confidence points from the centre heatmaps is extracted as the central position of the objects, and the position of the maximum confidence points from vertex heatmaps is extracted as the vertex of the objects.  However, the difference between the sizes of the output heatmaps and the input images is four times larger . Because the coordinate position can only be an integer, floating point deviation loss will occur in the mapping process, which affects the positioning of slender oriented objects. To compensate for the floating point bias loss, we predict the offset map $ O\in {R}^{4\times\frac{W}{d}\times\frac{H}{d}\times C} $. Given the peak point $ \widetilde{c}=(\widetilde{c_x},\widetilde{c_y}) $ and $ \widetilde{t}=(\widetilde{t_x},\widetilde{t_y}) $ in each input image, the floating point deviation value $ O $ is calculated using the following formula:
\begin{equation}
    O = [ \frac{\widetilde{c_x}}{d} - \lfloor\frac{\widetilde{c_x}}{d}\rfloor,\frac{\widetilde{c_y}}{d} - \lfloor\frac{\widetilde{c_y}}{d}\rfloor,\frac{\widetilde{t_x}}{d} - \lfloor\frac{\widetilde{t_x}}{d}\rfloor,\frac{\widetilde{t_y}}{d} - \lfloor\frac{\widetilde{t_y}}{d}\rfloor ]
\end{equation}
In this study, targets of the same category share one vertex float offset and one centre float offset. In the training stage, the offset is optimised with a smooth $ L_1 $ loss as follows.
\begin{equation}
    L_{offsets} = \frac{1}{N}\sum_{i=1}^NSmooth_{L_1}(O_i - \acute{O_i})
\end{equation}
\subsection{Framework}\label{subsec4}
Figure \ref{fig2} illustrates the overarching pipeline of the proposed ROTP. The network architecture is divided into three parts: the backbone, neck, and output heads. In addition, the formulation of the input image size is $ I\in {R}^{H \times W \times 3} $, where $ W $ and $ H $ are the width and height, respectively, of the input image. The input image is first sent to the backbone for the convolution operation to obtain feature maps’ these feature maps are then sent to the neck. In the present study, ResNet101\cite{he2016deep} was selected as the ROTP encoder-decoder. The neck is a deformed FPN structure. The FPN extracts different feature maps of the receptive field. To ensure that the detector detects large and small targets simultaneously, the feature maps are combined with up-sampling before being sent to the heads. ROTP outputs five feature maps of size $ F\in {R}^{\frac{W}{d}\times\frac{H}{d}\times S} $, where $ S $ is the degree of depth of the corresponding prediction of outputs, and $ d $ is the output step size, which was set to four in this study. Of the five output maps, one is the vertex predicted in the form of a heatmap $ (T\in {R}^{\frac{W}{d}\times\frac{H}{d}\times cls}) $, and the other$ (C\in {R}^{\frac{W}{d}\times\frac{H}{d}\times cls}) $ is the centre point predicted by the heatmap form; the other three are the length and width of the regression target $ (Reg\in {R}^{\frac{W}{d}\times\frac{H}{d}\times 2}) $, the offsets of the peaks$ (O\in {R}^{\frac{W}{d}\times\frac{H}{d}\times 4}) $, and the direction of the vertex $ (D\in {R}^{\frac{W}{d}\times\frac{H}{d}\times 1}) $ relative to the centre point.
\subsection{Loss Function}\label{subsec5}
In the training stage, the regress loss $ L_{Reg} $ of the width and height of the objects is also optimised with a smooth $ L_1 $ loss. The loss function of the proposed method is defined as follows:
\begin{equation}
L={\lambda_0L}_{ht}+\lambda_1L_{hc}+\lambda_2L_{Reg}+{\lambda_3L}_{offsets}+{\lambda_4L}_{D}
\end{equation}
$ \lambda_0 $, $ \lambda_1 $ , $ \lambda_2 $ , $ \lambda_3 $ , and $ \lambda_4 $ are the hyperparameters that balance all loss functions, where the inner $ \lambda_3 $ is 0.1, and the rest are set to 1.0; all losses except for $ L_{ht} $ and $ L_{hc} $ are calculated only when $ \rho^\prime $ is 1.0.

\section{Experiment}\label{sec4}
\subsection{Datasets}\label{subsec1}
The HRSC2016 dataset was used in the present study. In the annotation attribute of HRSC2016, there are annotations for vertex and centre point positions, and we directly use this information for training, whereas in the UCAS-AOD dataset, the marks in some images were not marked according to the four points clockwise from the top left corner of the bounding box, which misled our method. We marked the vertex first, then marked the centre point, and remarked these targets in terms of the long and short sides of the original objects. We used the DOTA datasets to verify the effectiveness of our method in various categories.

\noindent\textbf{HRSC2016}\cite{liu2016ship}
The dataset is a remote ship detection dataset, containing 1061 images marked with a rotating bounding box, ranging in size from $ 300 \times 300 $ to $ 1500 \times 900 $. The standard mAP evaluation protocol was used to evaluate the HRSC2016 dataset.

\noindent\textbf{UCAS-AOD}\cite{zhu2015orientation}
The dataset comprises 1000 aircraft images with 7482 instances and 510 car images with 7114 instances. The resolution of the UCASAOD is approximately $ 1280 \times 700 $, which is composed of a horizontal label box and a slanted label box. We used only the slanted label box for the training.

\noindent\textbf{DOTA}\cite{xia2018dota}
The dataset contains 2806 aerial images in 15 categories. In this dataset, the smallest image is $ 800 \times 800 $ and the largest is $ 4000 \times 4000 $. The training, validation, and test sets were divided into a $ 3:1:2 $ ratio. We cut the image to $ 1024 \times 1024 $ and the gap was set 200 pixels as the step size.
\subsection{Implementation Details}\label{subsec2}
In the training stage, ROTP used Resnet101\cite{he2016deep} as the backbone, the image input resolution is set to $ 800 \times 800 $ , and the size of the output feature map is $ 200\times200 $ . The image sizes of the training data differed in the present study. Ensuring that the target would not be deformed during the process of image scaling, we scaled the image without changing its aspect ratio, and zero filling was carried out for the rest. For the UCASAOD dataset, $ 70\% $ of the data were shuffled as the training set and $ 30\% $ as the test set. For the DOTA dataset, direct scaling would cause object loss owing to the high image resolution; we therefore cut the DOTA image to a $ 1024\times1024 $ size and gap set to 200. For all the data, we used random rotation, random flip, pixel transformation, and other processing methods to enhance the data. Adam\cite{kingma2014adam} acted as an optimiser for our network and uses a warmup policy, with batch size set to 4, and training to loss convergence. In the testing stage, the test images were scaled in the same way as the train images scaled in the training stage. The first 200 key points were selected, and the confidence was set to 0.25. For other datasets, we calculated AP with the default IOU parameter in PASCAL VOC\cite{everingham2010pascal}, that is, 0.5. The accuracy is used for performance evaluation, which is well sued in other data mining and machine learning tasks \cite{IJCAT2012,6729567}. All experiments were trained with Pytorch on an Nvidia RTX 2080TI GPU.
\begin{figure}[H]%
    \centering
    \includegraphics[width=0.8\textwidth]{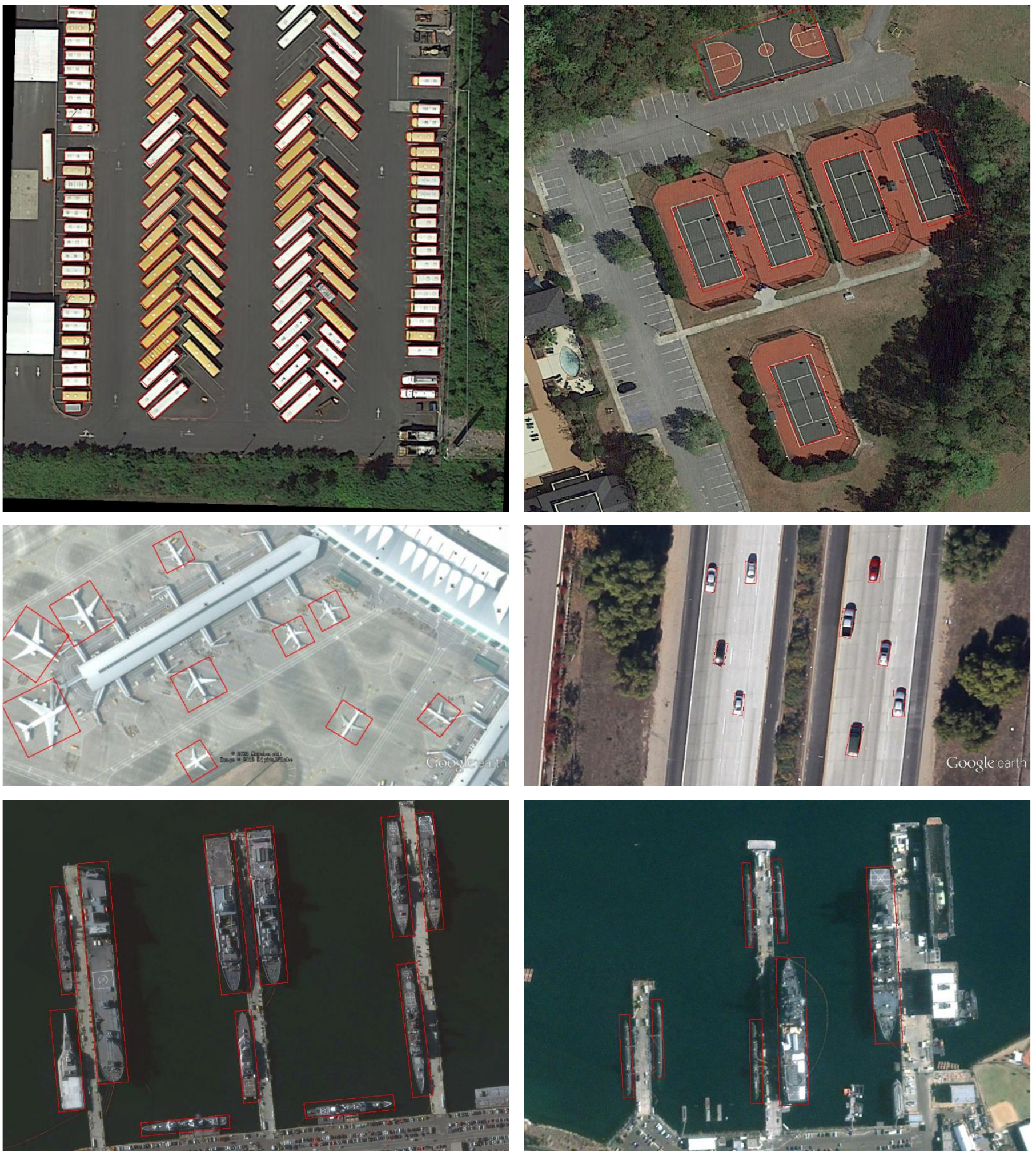}
    \caption{Visualization of the detection results of DOTA, UCAS-AOD and HRSC2016.
}
\label{fig6}
\end{figure}

\subsection{Comparisons with STATE-OF-THE-ART}\label{subsec3}
\begin{table}[h]
\begin{center}
\caption{Comparisons on DOTA between ROPT and some classcial oriented detection methods.}\label{tab1}%
\resizebox{\textwidth}{!}{
\begin{tabular}{@{}lllllllllllllllll@{}}
\toprule
Method & Pl	& Bd &Br	&Gft	&Sv	&Lv	&Sh	&Tc	&Bc	&St	&Sbf	&Ra	&Ha	&Sp 	&He 	&mAP \\
\midrule
R2CNN\cite{jiang2017r2cnn}  &80.94  &65.67 &35.34 &67.44 &59.92 &50.91 &55.81 &90.67 &66.92 &72.39 &55.06 &52.23 &55.14 &53.35 &48.22 &60.67  \\
R2CNN\cite{jiang2017r2cnn}  &80.94  &65.67 &35.34 &67.44 &59.92 &50.91 &55.81 &90.67 &66.92 &72.39 &55.06 &52.23 &55.14 &53.35 &48.22 &60.67  \\
ICN\cite{azimi2018towards}    &81.40  &74.30 &47.70 &70.30 &64.90 &67.80 &70.00 &90.80 &79.10 &78.20 &53.60 &62.90 &67.00 &64.20 &50.20 &68.20  \\
RRPN\cite{ma2018arbitrary}  &88.52 &71.20 &31.66 &59.30 &51.85 &56.19 &57.25 &90.81 &72.84 &67.38 &56.69 &52.84 &53.08 &51.94 &53.58 &61.01 \\
R-DFPN\cite{yang2018automatic}  &80.92  &65.82 &33.77 &58.94 &55.77& 50.94 &54.78 &90.33 &66.34 &68.66 &48.73 &51.76 &55.10 &51.32 &35.88 &57.94    \\
RoI-Transformer\cite{ding2018learning}  &\textbf{88.64} &78.52 &43.44 &\textbf{75.92} &68.81 &73.68 &\textbf{83.59} &90.74 &77.27 &81.46 &\textbf{58.39} &53.54 &62.83 &58.93 &47.67 &69.56                    \\
P-RSDet\cite{zhou2020arbitrary}  &88.58  &77.84 &\textbf{50.44} &69.29 &71.10 &\textbf{75.79} &78.66 &\textbf{90.88} &80.10 &81.71 &57.92 &\textbf{63.03} &66.30 &\textbf{69.77} &\textbf{63.13} &\textbf{72.30}                   \\
ROPT    &81.17  &\textbf{84.34} &47.33 &62.65 &\textbf{71.42} &71.04 &78.09 &89.93 &\textbf{80.34} &\textbf{84.78} &44.74 &60.81 &\textbf{66.43} &69.14 &62.15 &70.29            \\
\botrule
\end{tabular}}

\end{center}
\end{table}
The results of the comparison of the performance of ROTP and some famous one-stage or two-stage oriented object detection methods on DOTA are illustrated in Table \ref{tab1}. The accuracy of our method for Bd, Sv, Bc,St, and Ha was $ 5.82\% $,$ 0.32\% $, $ 0.24\% $ , $ 3.07\% $ and $ 0.13\% $ higher, respectively, than those of the second method. The total mAP was $ 2.01\% $ lower than that of P-RSDET which is also based on the anchor free heatmap mothed. This is because the AP value of our model was very low when detecting Br and Sbf categories, resulting in a decline in overall accuracy. In addition, the accuracy of the category is competitive with that of other methods.
\begin{table}[h]
\begin{center}
\begin{minipage}{174pt}
\caption{meanAveragePrecision (\%). Comparison results between ROTP and other classical oriented object detectors on UCAS-AOD. During calculating mAP, we set IOU in 0.5 and confidence in 0.25.}\label{tab2}%
\begin{tabular}{@{}llll@{}}
\toprule
Method & Plane	& Car  	&mAP \\
\midrule
RRPN\cite{ma2018arbitrary} & 88.04 & 74.36 & 81.2\\
$ R^2CNN $\cite{jiang2017r2cnn}       &89.76 	&78.89 	&84.32        \\
R-DFPN\cite{yang2018automatic}        &88.91	&81.27	&85.09        \\
X-LineNet\cite{wei2020x}       &91.3	&-	&-         \\
P-RSDet\cite{zhou2020arbitrary}   &92.69 	&\textbf{87.38}	&90.03    \\
ROPT        &\textbf{95.42}	&84.76	&\textbf{90.09} \\
\botrule
\end{tabular}

\end{minipage}
\end{center}
\end{table}

The results of the comparison of the performance of ROTP and other classical-oriented object detectors on the UCAS-AOD dataset are shown in Table \ref{tab2}. ROTP demonstrated a certain degree of improvement over other methods in mAP. P-RSDet also used a single-stage heatmap method to detect rotating targets and achieved a $ 90.3\% $mAP; ROTP was $ 2.73\% $ higher than P-RSDET in the aircraft category, but $ 2.62\% $ lower in the car category. We assume this to be because P-RSDET cuts the image to $ 512\times512 $ , while we scaled the image at equal proportions. Our method will make the small target size smaller, reducing the detection accuracy of the automobile category.

\begin{table}[h]
\begin{center}
\caption{Comparisons on HRSC2016 with oriented bounding boxes. Work used the VOC07 method to calculate mAP(\%)}\label{tab3}%
\begin{tabular}{@{}lllllll@{}}
\toprule
Method    & $ R^2CNN $\cite{jiang2017r2cnn}	  & RC1\&RC2\cite{inproceedings}	  & Axis Learning\cite{xiao2020axis}  & RRPN\cite{ma2018arbitrary}	  &TOSO\cite{feng2020toso}	&ROTP \\
\midrule
mAP    & 73.03 	       & 75.7 	 & 78.15            & 79.08    & 79.29    & \textbf{80.6}   \\
\botrule
\end{tabular}

\end{center}
\end{table}

The results of the comparison of performance for HRSC2016 of ROTP and other mainstream deep-learning-based methods are shown in Table      \ref{tab3}. R2CNN adds multi-scale ROI pooling and inclined frame detection based on FastRCNN, achieving an AP of 73.07. The RRPN introduced several tilted anchors and reached 79.08 AP, while our model reached 80.6 AP, representing a certain improvement over the above algorithm. However, shortcomings are also present that require many data enhancement strategies to ensure the uptake of positive samples.
\subsection{Ablation Studies}\label{subsec3}
We examine the availability of the proposed method from three perspectives: the separate use of angles or angles with key points, different encoder-decoders, and different heatmap generation methods.
\begin{table}[h]
\begin{center}
\begin{minipage}{174pt}
\caption{Comparison on different Encoder-Decoder.}\label{tab4}%
\begin{tabular}{@{}llll@{}}
\toprule
Encoder-Decoder & Plane	& Car  	&mAP \\
\midrule
Unet\cite{2015U}        & 93.69	&86.15	&89.92\\
104-Hourglass\cite{2016Stacked} & 96.34&88.56        \\
ROPT         &95.42	&84.76	&90.09 \\
\botrule
\end{tabular}

\end{minipage}
\end{center}
\end{table}

\noindent\textbf{Different encoders and decoders:} 
In ROTP, we used the improved Restnet101 with FPN as the encoders and decoders, as shown in Table \ref{tab4}. To test the impact of different encoders and decoders on our model, we replaced the previous Heads architecture with the network architecture of Unet and 104-Hourglass. We conducted experiments using the UCASAOD dataset. As shown in the table, the AP of our codec is $ 1.73\% $ higher than that of Unet in the aircraft category, but that of Unet is $ 1.43\% $ higher than that of our codec in the automobile category. This result is caused by the network structure. To keep the ROTP robust in the detection of targets at different scales, the network structure is built by combining different feature maps from the FPN; this is well reflected in the airplane image. However, in the car image, the change in the size of the car target is small, which can be understood to be a small target. After resizing, the car image becomes smaller, and a feature image with a large receptive field will lose information for the small target, which will cause little interference to the feature image after combination. However, unit-smooth convolution processing does not involve this shortcoming. In addition, using hourglass as the backbone improves the mAP by $ 2.36\% $ over that of Resnet101. The experiments show that the proposed improved network structure is effective.

\begin{figure}[H]
    \includegraphics[scale=0.6]{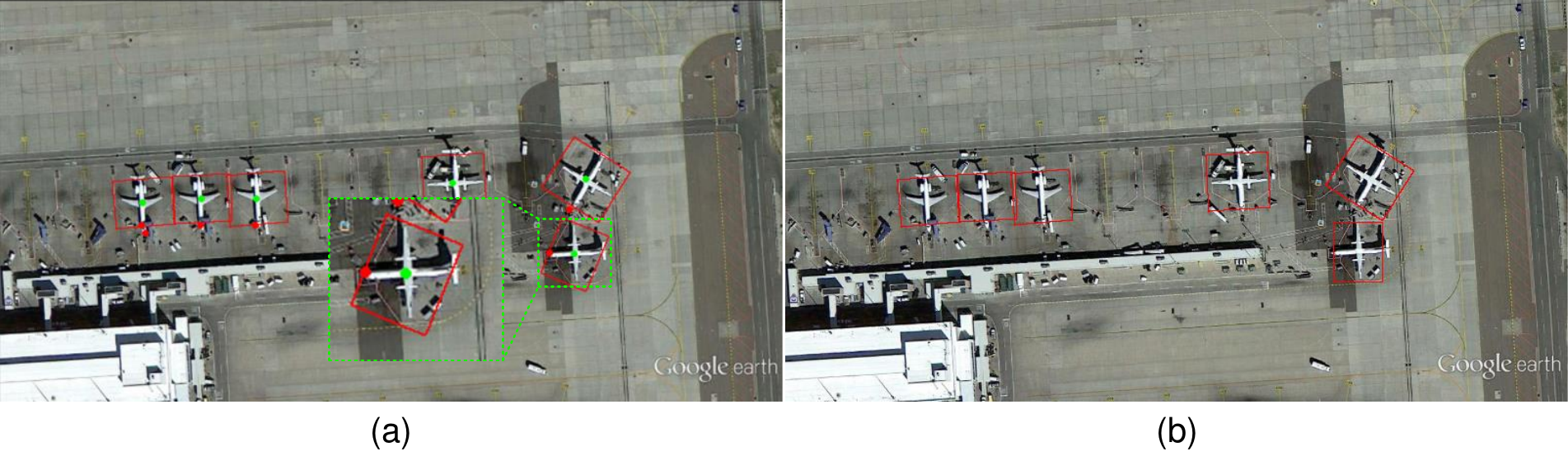}
    \centering
    \caption{Comparison of KeyPoints Match.(a) the direction is predicted without matching the corresponding vertices;(b) predict direction and match vertices}
    \label{fig7}
\end{figure}

\begin{table}[h]
\begin{center}
\begin{minipage}{174pt}
\caption{Comparison between angles and keypoints with angles}\label{tab5}%
\begin{tabular}{@{}llll@{}}
\toprule
KeyPoints Match & Plane	& Car  	&mAP \\
\midrule
Angles        & 93.69	&86.15	&89.92\\
Angles and Keypoints &95.42	&84.76	&90.09\\
\botrule
\end{tabular}

\end{minipage}
\end{center}
\end{table}

\noindent\textbf{KeyPoints Match:} 
In this study, we added the top prediction to the direction prediction. To demonstrate the advanced nature of our proposed method, various experiments were carried out on the UCAS-AOD. As shown in Table \ref{tab5}, when the keypoint matching method was adopted, the mAP value was $ 8.97\% $ higher than that of the angle prediction method alone. As shown in Figure \ref{fig7}, the direction prediction was not accurate for some objects, leading to failure of prediction. Therefore, the keypoint matching method is effective for ROTP.
\begin{table}[h]
\begin{center}
\begin{minipage}{174pt}
\caption{Comparison between Gauss Heatmap and Ours}\label{tab6}%
\begin{tabular}{@{}llll@{}}
\toprule
Heatmap Generation & Plane	& Car  	&mAP \\
\midrule
Gauss Heatmap        &92.26	&79.62	&85.94\\
Ours &95.42	&84.76	&90.09\\
\botrule
\end{tabular}

\end{minipage}
\end{center}
\end{table}
\noindent\textbf{Difference in Heatmaps:}
A comparison of the Solar Coron and Gaussian heatmaps is shown in Table \ref{tab6}. In the comparison experiment, a directed target detection experiment was carried out on the UCASAOD dataset. based on different heatmap generation schemes. Owing to the high aspect ratio and the differing sizes of the directed targets, Gaussian heatmaps have a low AP when treating small targets, as shown in Figure \ref{fig4}. It may also be that the original small target becomes smaller upon resizing the input images. Finally, using the Solar Corona heatmap improves the mAP by $ 4.15\% $ over that of the Gaussian heatmap, thus proving the effectiveness of our heatmap generation method.

\section{Conclusion}\label{sec5}
In the present study, an instance generation method is proposed for rotating target representation and solar corona heatmap (SCH) generation for multi-object-oriented perception. A single-stage rotating object detector via two points with a solar corona heatmap ROTP was proposed for rotating object detection. Via simultaneous vertex and centre point detection, ROTP can avoid the position offset caused by the direct prediction angle, and the model can achieve good results without complex pre-design without using anchors and can detect key points. This method enables ROTP to quickly detect densely arranged objects of different sizes , such as cars, boats, and small planes. Experimental results on multiple datasets indicate that the modelling of rotation detectors based on vertices and centre points is effective. However, this study also has shortcomings. If the central area of the target to be detected is covered, ROTP will not be able to capture the real central position of the instance, resulting in detection failure. In addition, our attempted to use ROTP to detect scene text images\cite{2015ICDAR2015,karatzas2013icdar} did not have a satisfactory result. The real centre point of a text sentence cannot be captured based on a single central heatmap. In fact, detection targeting of a single character is the correct way to use a single-stage heatmap to detect scene text. In future work, researchers are encouraged to mark the vertices of the rotating target or other useful key points, and we hope to extend this work to the field of 3D detection and tracking.

\bibliography{sn-bibliography}


\begin{thebibliography}{46}
\ifx \bisbn   \undefined \def \bisbn  #1{ISBN #1}\fi
\ifx \binits  \undefined \def \binits#1{#1}\fi
\ifx \bauthor  \undefined \def \bauthor#1{#1}\fi
\ifx \batitle  \undefined \def \batitle#1{#1}\fi
\ifx \bjtitle  \undefined \def \bjtitle#1{#1}\fi
\ifx \bvolume  \undefined \def \bvolume#1{\textbf{#1}}\fi
\ifx \byear  \undefined \def \byear#1{#1}\fi
\ifx \bissue  \undefined \def \bissue#1{#1}\fi
\ifx \bfpage  \undefined \def \bfpage#1{#1}\fi
\ifx \blpage  \undefined \def \blpage #1{#1}\fi
\ifx \burl  \undefined \def \burl#1{\textsf{#1}}\fi
\ifx \doiurl  \undefined \def \doiurl#1{\url{https://doi.org/#1}}\fi
\ifx \betal  \undefined \def \betal{\textit{et al.}}\fi
\ifx \binstitute  \undefined \def \binstitute#1{#1}\fi
\ifx \binstitutionaled  \undefined \def \binstitutionaled#1{#1}\fi
\ifx \bctitle  \undefined \def \bctitle#1{#1}\fi
\ifx \beditor  \undefined \def \beditor#1{#1}\fi
\ifx \bpublisher  \undefined \def \bpublisher#1{#1}\fi
\ifx \bbtitle  \undefined \def \bbtitle#1{#1}\fi
\ifx \bedition  \undefined \def \bedition#1{#1}\fi
\ifx \bseriesno  \undefined \def \bseriesno#1{#1}\fi
\ifx \blocation  \undefined \def \blocation#1{#1}\fi
\ifx \bsertitle  \undefined \def \bsertitle#1{#1}\fi
\ifx \bsnm \undefined \def \bsnm#1{#1}\fi
\ifx \bsuffix \undefined \def \bsuffix#1{#1}\fi
\ifx \bparticle \undefined \def \bparticle#1{#1}\fi
\ifx \barticle \undefined \def \barticle#1{#1}\fi
\bibcommenthead
\ifx \bconfdate \undefined \def \bconfdate #1{#1}\fi
\ifx \botherref \undefined \def \botherref #1{#1}\fi
\ifx \url \undefined \def \url#1{\textsf{#1}}\fi
\ifx \bchapter \undefined \def \bchapter#1{#1}\fi
\ifx \bbook \undefined \def \bbook#1{#1}\fi
\ifx \bcomment \undefined \def \bcomment#1{#1}\fi
\ifx \oauthor \undefined \def \oauthor#1{#1}\fi
\ifx \citeauthoryear \undefined \def \citeauthoryear#1{#1}\fi
\ifx \endbibitem  \undefined \def \endbibitem {}\fi
\ifx \bconflocation  \undefined \def \bconflocation#1{#1}\fi
\ifx \arxivurl  \undefined \def \arxivurl#1{\textsf{#1}}\fi
\csname PreBibitemsHook\endcsname

\bibitem{redmon2016you}
\begin{bchapter}
\bauthor{\bsnm{Redmon}, \binits{J.}},
\bauthor{\bsnm{Divvala}, \binits{S.}},
\bauthor{\bsnm{Girshick}, \binits{R.}},
\bauthor{\bsnm{Farhadi}, \binits{A.}}:
\bctitle{You only look once: Unified, real-time object detection}.
In: \bbtitle{Proceedings of the IEEE Conference on Computer Vision and Pattern
  Recognition},
pp. \bfpage{779}--\blpage{788}
(\byear{2016})
\end{bchapter}
\endbibitem

\bibitem{liu2016ssd}
\begin{bchapter}
\bauthor{\bsnm{Liu}, \binits{W.}},
\bauthor{\bsnm{Anguelov}, \binits{D.}},
\bauthor{\bsnm{Erhan}, \binits{D.}},
\bauthor{\bsnm{Szegedy}, \binits{C.}},
\bauthor{\bsnm{Reed}, \binits{S.}},
\bauthor{\bsnm{Fu}, \binits{C.-Y.}},
\bauthor{\bsnm{Berg}, \binits{A.C.}}:
\bctitle{Ssd: Single shot multibox detector}.
In: \bbtitle{European Conference on Computer Vision},
pp. \bfpage{21}--\blpage{37}
(\byear{2016}).
\bcomment{Springer}
\end{bchapter}
\endbibitem

\bibitem{girshick2015fast}
\begin{bchapter}
\bauthor{\bsnm{Girshick}, \binits{R.}}:
\bctitle{Fast r-cnn}.
In: \bbtitle{Proceedings of the IEEE International Conference on Computer
  Vision},
pp. \bfpage{1440}--\blpage{1448}
(\byear{2015})
\end{bchapter}
\endbibitem

\bibitem{jiang2017r2cnn}
\begin{botherref}
\oauthor{\bsnm{Jiang}, \binits{Y.}},
\oauthor{\bsnm{Zhu}, \binits{X.}},
\oauthor{\bsnm{Wang}, \binits{X.}},
\oauthor{\bsnm{Yang}, \binits{S.}},
\oauthor{\bsnm{Li}, \binits{W.}},
\oauthor{\bsnm{Wang}, \binits{H.}},
\oauthor{\bsnm{Fu}, \binits{P.}},
\oauthor{\bsnm{Luo}, \binits{Z.}}:
R2cnn: rotational region cnn for orientation robust scene text detection.
arXiv preprint arXiv:1706.09579
(2017)
\end{botherref}
\endbibitem

\bibitem{ma2018arbitrary}
\begin{barticle}
\bauthor{\bsnm{Ma}, \binits{J.}},
\bauthor{\bsnm{Shao}, \binits{W.}},
\bauthor{\bsnm{Ye}, \binits{H.}},
\bauthor{\bsnm{Wang}, \binits{L.}},
\bauthor{\bsnm{Wang}, \binits{H.}},
\bauthor{\bsnm{Zheng}, \binits{Y.}},
\bauthor{\bsnm{Xue}, \binits{X.}}:
\batitle{Arbitrary-oriented scene text detection via rotation proposals}.
\bjtitle{IEEE Transactions on Multimedia}
\bvolume{20}(\bissue{11}),
\bfpage{3111}--\blpage{3122}
(\byear{2018})
\end{barticle}
\endbibitem

\bibitem{liao2017textboxes}
\begin{bchapter}
\bauthor{\bsnm{Liao}, \binits{M.}},
\bauthor{\bsnm{Shi}, \binits{B.}},
\bauthor{\bsnm{Bai}, \binits{X.}},
\bauthor{\bsnm{Wang}, \binits{X.}},
\bauthor{\bsnm{Liu}, \binits{W.}}:
\bctitle{Textboxes: A fast text detector with a single deep neural network}.
In: \bbtitle{Thirty-first AAAI Conference on Artificial Intelligence}
(\byear{2017})
\end{bchapter}
\endbibitem

\bibitem{liao2018textboxes++}
\begin{barticle}
\bauthor{\bsnm{Liao}, \binits{M.}},
\bauthor{\bsnm{Shi}, \binits{B.}},
\bauthor{\bsnm{Bai}, \binits{X.}}:
\batitle{Textboxes++: A single-shot oriented scene text detector}.
\bjtitle{IEEE transactions on image processing}
\bvolume{27}(\bissue{8}),
\bfpage{3676}--\blpage{3690}
(\byear{2018})
\end{barticle}
\endbibitem

\bibitem{zhou2020arbitrary}
\begin{barticle}
\bauthor{\bsnm{Zhou}, \binits{L.}},
\bauthor{\bsnm{Wei}, \binits{H.}},
\bauthor{\bsnm{Li}, \binits{H.}},
\bauthor{\bsnm{Zhao}, \binits{W.}},
\bauthor{\bsnm{Zhang}, \binits{Y.}},
\bauthor{\bsnm{Zhang}, \binits{Y.}}:
\batitle{Arbitrary-oriented object detection in remote sensing images based on
  polar coordinates}.
\bjtitle{IEEE Access}
\bvolume{8},
\bfpage{223373}--\blpage{223384}
(\byear{2020})
\end{barticle}
\endbibitem

\bibitem{2021Oriented}
\begin{bchapter}
\bauthor{\bsnm{Yi}, \binits{J.}},
\bauthor{\bsnm{Wu}, \binits{P.}},
\bauthor{\bsnm{Liu}, \binits{B.}},
\bauthor{\bsnm{Huang}, \binits{Q.}},
\bauthor{\bsnm{Metaxas}, \binits{D.}}:
\bctitle{Oriented object detection in aerial images with box boundary-aware
  vectors}.
In: \bbtitle{2021 IEEE Winter Conference on Applications of Computer Vision
  (WACV)}
(\byear{2021})
\end{bchapter}
\endbibitem

\bibitem{xu2020gliding}
\begin{barticle}
\bauthor{\bsnm{Xu}, \binits{Y.}},
\bauthor{\bsnm{Fu}, \binits{M.}},
\bauthor{\bsnm{Wang}, \binits{Q.}},
\bauthor{\bsnm{Wang}, \binits{Y.}},
\bauthor{\bsnm{Chen}, \binits{K.}},
\bauthor{\bsnm{Xia}, \binits{G.-S.}},
\bauthor{\bsnm{Bai}, \binits{X.}}:
\batitle{Gliding vertex on the horizontal bounding box for multi-oriented
  object detection}.
\bjtitle{IEEE transactions on pattern analysis and machine intelligence}
\bvolume{43}(\bissue{4}),
\bfpage{1452}--\blpage{1459}
(\byear{2020})
\end{barticle}
\endbibitem

\bibitem{redmon2018yolov3}
\begin{botherref}
\oauthor{\bsnm{Redmon}, \binits{J.}},
\oauthor{\bsnm{Farhadi}, \binits{A.}}:
Yolov3: An incremental improvement.
arXiv preprint arXiv:1804.02767
(2018)
\end{botherref}
\endbibitem

\bibitem{zhou2017east}
\begin{bchapter}
\bauthor{\bsnm{Zhou}, \binits{X.}},
\bauthor{\bsnm{Yao}, \binits{C.}},
\bauthor{\bsnm{Wen}, \binits{H.}},
\bauthor{\bsnm{Wang}, \binits{Y.}},
\bauthor{\bsnm{Zhou}, \binits{S.}},
\bauthor{\bsnm{He}, \binits{W.}},
\bauthor{\bsnm{Liang}, \binits{J.}}:
\bctitle{East: an efficient and accurate scene text detector}.
In: \bbtitle{Proceedings of the IEEE Conference on Computer Vision and Pattern
  Recognition},
pp. \bfpage{5551}--\blpage{5560}
(\byear{2017})
\end{bchapter}
\endbibitem

\bibitem{DBLPXing}
\begin{botherref}
\oauthor{\bsnm{Su}, \binits{X.}},
\oauthor{\bsnm{Xue}, \binits{S.}},
\oauthor{\bsnm{Liu}, \binits{F.}},
\oauthor{\bsnm{Wu}, \binits{J.}},
\oauthor{\bsnm{Yang}, \binits{J.}},
\oauthor{\bsnm{Zhou}, \binits{C.}},
\oauthor{\bsnm{Hu}, \binits{W.}},
\oauthor{\bsnm{Paris}, \binits{C.}},
\oauthor{\bsnm{Nepal}, \binits{S.}},
\oauthor{\bsnm{Jin}, \binits{D.}},
\oauthor{\bsnm{Sheng}, \binits{Q.Z.}},
\oauthor{\bsnm{Yu}, \binits{P.S.}}:
A comprehensive survey on community detection with deep learning.
CoRR
\textbf{abs/2105.12584}
(2021)
\end{botherref}
\endbibitem

\bibitem{9565320}
\begin{botherref}
\oauthor{\bsnm{Ma}, \binits{X.}},
\oauthor{\bsnm{Wu}, \binits{J.}},
\oauthor{\bsnm{Xue}, \binits{S.}},
\oauthor{\bsnm{Yang}, \binits{J.}},
\oauthor{\bsnm{Zhou}, \binits{C.}},
\oauthor{\bsnm{Sheng}, \binits{Q.Z.}},
\oauthor{\bsnm{Xiong}, \binits{H.}},
\oauthor{\bsnm{Akoglu}, \binits{L.}}:
A comprehensive survey on graph anomaly detection with deep learning.
IEEE Transactions on Knowledge and Data Engineering,
1--1
(2021).
\doiurl{10.1109/TKDE.2021.3118815}
\end{botherref}
\endbibitem

\bibitem{IJCAI20Fanzhen}
\begin{bchapter}
\bauthor{\bsnm{Liu}, \binits{F.}},
\bauthor{\bsnm{Xue}, \binits{S.}},
\bauthor{\bsnm{Wu}, \binits{J.}},
\bauthor{\bsnm{Zhou}, \binits{C.}},
\bauthor{\bsnm{Hu}, \binits{W.}},
\bauthor{\bsnm{Paris}, \binits{C.}},
\bauthor{\bsnm{Nepal}, \binits{S.}},
\bauthor{\bsnm{Yang}, \binits{J.}},
\bauthor{\bsnm{Yu}, \binits{P.S.}}:
\bctitle{Deep learning for community detection: Progress, challenges and
  opportunities}.
In: \bbtitle{Proceedings of the Twenty-Ninth International Joint Conference on
  Artificial Intelligence}.
\bsertitle{IJCAI'20}
(\byear{2021})
\end{bchapter}
\endbibitem

\bibitem{2014}
\begin{botherref}
\oauthor{\bsnm{Girshick}, \binits{R.}},
\oauthor{\bsnm{Donahue}, \binits{J.}},
\oauthor{\bsnm{Darrell}, \binits{T.}},
\oauthor{\bsnm{Malik}, \binits{J.}}:
[ieee 2014 ieee conference on computer vision and pattern recognition (cvpr) -
  columbus, oh, usa (2014.6.23-2014.6.28)] 2014 ieee conference on computer
  vision and pattern recognition - rich feature hierarchies for accurate object
  detection and semantic se,
580--587
(2014)
\end{botherref}
\endbibitem

\bibitem{2016Faster}
\begin{bchapter}
\bauthor{\bsnm{Ren}, \binits{S.}},
\bauthor{\bsnm{He}, \binits{K.}},
\bauthor{\bsnm{Girshick}, \binits{R.}},
\bauthor{\bsnm{Sun}, \binits{J.}}:
\bctitle{Faster r-cnn: Towards real-time object detection with region proposal
  networks}.
In: \bbtitle{NIPS}
(\byear{2016})
\end{bchapter}
\endbibitem

\bibitem{redmon2017yolo9000}
\begin{bchapter}
\bauthor{\bsnm{Redmon}, \binits{J.}},
\bauthor{\bsnm{Farhadi}, \binits{A.}}:
\bctitle{Yolo9000: better, faster, stronger}.
In: \bbtitle{Proceedings of the IEEE Conference on Computer Vision and Pattern
  Recognition},
pp. \bfpage{7263}--\blpage{7271}
(\byear{2017})
\end{bchapter}
\endbibitem

\bibitem{lin2017focal}
\begin{bchapter}
\bauthor{\bsnm{Lin}, \binits{T.-Y.}},
\bauthor{\bsnm{Goyal}, \binits{P.}},
\bauthor{\bsnm{Girshick}, \binits{R.}},
\bauthor{\bsnm{He}, \binits{K.}},
\bauthor{\bsnm{Doll{\'a}r}, \binits{P.}}:
\bctitle{Focal loss for dense object detection}.
In: \bbtitle{Proceedings of the IEEE International Conference on Computer
  Vision},
pp. \bfpage{2980}--\blpage{2988}
(\byear{2017})
\end{bchapter}
\endbibitem

\bibitem{2017DSSD}
\begin{botherref}
\oauthor{\bsnm{Fu}, \binits{C.Y.}},
\oauthor{\bsnm{Liu}, \binits{W.}},
\oauthor{\bsnm{Ranga}, \binits{A.}},
\oauthor{\bsnm{Tyagi}, \binits{A.}},
\oauthor{\bsnm{Berg}, \binits{A.C.}}:
Dssd : Deconvolutional single shot detector
(2017)
\end{botherref}
\endbibitem

\bibitem{2019FoveaBox}
\begin{botherref}
\oauthor{\bsnm{Kong}, \binits{T.}},
\oauthor{\bsnm{Sun}, \binits{F.}},
\oauthor{\bsnm{Liu}, \binits{H.}},
\oauthor{\bsnm{Jiang}, \binits{Y.}},
\oauthor{\bsnm{Shi}, \binits{J.}}:
Foveabox: Beyond anchor-based object detector
(2019)
\end{botherref}
\endbibitem

\bibitem{2019RepPoints}
\begin{botherref}
\oauthor{\bsnm{Yang}, \binits{Z.}},
\oauthor{\bsnm{Liu}, \binits{S.}},
\oauthor{\bsnm{Hu}, \binits{H.}},
\oauthor{\bsnm{Wang}, \binits{L.}},
\oauthor{\bsnm{Lin}, \binits{S.}}:
Reppoints: Point set representation for object detection.
IEEE
(2019)
\end{botherref}
\endbibitem

\bibitem{tian2019fcos}
\begin{bchapter}
\bauthor{\bsnm{Tian}, \binits{Z.}},
\bauthor{\bsnm{Shen}, \binits{C.}},
\bauthor{\bsnm{Chen}, \binits{H.}},
\bauthor{\bsnm{He}, \binits{T.}}:
\bctitle{Fcos: Fully convolutional one-stage object detection}.
In: \bbtitle{Proceedings of the IEEE/CVF International Conference on Computer
  Vision},
pp. \bfpage{9627}--\blpage{9636}
(\byear{2019})
\end{bchapter}
\endbibitem

\bibitem{law2018cornernet}
\begin{bchapter}
\bauthor{\bsnm{Law}, \binits{H.}},
\bauthor{\bsnm{Deng}, \binits{J.}}:
\bctitle{Cornernet: Detecting objects as paired keypoints}.
In: \bbtitle{Proceedings of the European Conference on Computer Vision (ECCV)},
pp. \bfpage{734}--\blpage{750}
(\byear{2018})
\end{bchapter}
\endbibitem

\bibitem{zhou2019objects}
\begin{botherref}
\oauthor{\bsnm{Zhou}, \binits{X.}},
\oauthor{\bsnm{Wang}, \binits{D.}},
\oauthor{\bsnm{Kr{\"a}henb{\"u}hl}, \binits{P.}}:
Objects as points.
arXiv preprint arXiv:1904.07850
(2019)
\end{botherref}
\endbibitem

\bibitem{2018Rotation}
\begin{botherref}
\oauthor{\bsnm{Liao}, \binits{M.}},
\oauthor{\bsnm{Zhu}, \binits{Z.}},
\oauthor{\bsnm{Shi}, \binits{B.}},
\oauthor{\bsnm{Xia}, \binits{G.S.}},
\oauthor{\bsnm{Bai}, \binits{X.}}:
Rotation-sensitive regression for oriented scene text detection.
IEEE
(2018)
\end{botherref}
\endbibitem

\bibitem{2015U}
\begin{botherref}
\oauthor{\bsnm{Ronneberger}, \binits{O.}},
\oauthor{\bsnm{Fischer}, \binits{P.}},
\oauthor{\bsnm{Brox}, \binits{T.}}:
U-net: Convolutional networks for biomedical image segmentation.
Springer International Publishing
(2015)
\end{botherref}
\endbibitem

\bibitem{ding2018learning}
\begin{botherref}
\oauthor{\bsnm{Ding}, \binits{J.}},
\oauthor{\bsnm{Xue}, \binits{N.}},
\oauthor{\bsnm{Long}, \binits{Y.}},
\oauthor{\bsnm{Xia}, \binits{G.-S.}},
\oauthor{\bsnm{Lu}, \binits{Q.}}:
Learning roi transformer for detecting oriented objects in aerial images.
arXiv preprint arXiv:1812.00155
(2018)
\end{botherref}
\endbibitem

\bibitem{azimi2018towards}
\begin{bchapter}
\bauthor{\bsnm{Azimi}, \binits{S.M.}},
\bauthor{\bsnm{Vig}, \binits{E.}},
\bauthor{\bsnm{Bahmanyar}, \binits{R.}},
\bauthor{\bsnm{K{\"o}rner}, \binits{M.}},
\bauthor{\bsnm{Reinartz}, \binits{P.}}:
\bctitle{Towards multi-class object detection in unconstrained remote sensing
  imagery}.
In: \bbtitle{Asian Conference on Computer Vision},
pp. \bfpage{150}--\blpage{165}
(\byear{2018}).
\bcomment{Springer}
\end{bchapter}
\endbibitem

\bibitem{2019Learning}
\begin{botherref}
\oauthor{\bsnm{Qian}, \binits{W.}},
\oauthor{\bsnm{Yang}, \binits{X.}},
\oauthor{\bsnm{Peng}, \binits{S.}},
\oauthor{\bsnm{Guo}, \binits{Y.}},
\oauthor{\bsnm{Yan}, \binits{J.}}:
Learning modulated loss for rotated object detection
(2019)
\end{botherref}
\endbibitem

\bibitem{he2016deep}
\begin{bchapter}
\bauthor{\bsnm{He}, \binits{K.}},
\bauthor{\bsnm{Zhang}, \binits{X.}},
\bauthor{\bsnm{Ren}, \binits{S.}},
\bauthor{\bsnm{Sun}, \binits{J.}}:
\bctitle{Deep residual learning for image recognition}.
In: \bbtitle{Proceedings of the IEEE Conference on Computer Vision and Pattern
  Recognition},
pp. \bfpage{770}--\blpage{778}
(\byear{2016})
\end{bchapter}
\endbibitem

\bibitem{liu2016ship}
\begin{barticle}
\bauthor{\bsnm{Liu}, \binits{Z.}},
\bauthor{\bsnm{Wang}, \binits{H.}},
\bauthor{\bsnm{Weng}, \binits{L.}},
\bauthor{\bsnm{Yang}, \binits{Y.}}:
\batitle{Ship rotated bounding box space for ship extraction from
  high-resolution optical satellite images with complex backgrounds}.
\bjtitle{IEEE Geoscience and Remote Sensing Letters}
\bvolume{13}(\bissue{8}),
\bfpage{1074}--\blpage{1078}
(\byear{2016})
\end{barticle}
\endbibitem

\bibitem{zhu2015orientation}
\begin{bchapter}
\bauthor{\bsnm{Zhu}, \binits{H.}},
\bauthor{\bsnm{Chen}, \binits{X.}},
\bauthor{\bsnm{Dai}, \binits{W.}},
\bauthor{\bsnm{Fu}, \binits{K.}},
\bauthor{\bsnm{Ye}, \binits{Q.}},
\bauthor{\bsnm{Jiao}, \binits{J.}}:
\bctitle{Orientation robust object detection in aerial images using deep
  convolutional neural network}.
In: \bbtitle{2015 IEEE International Conference on Image Processing (ICIP)},
pp. \bfpage{3735}--\blpage{3739}
(\byear{2015}).
\bcomment{IEEE}
\end{bchapter}
\endbibitem

\bibitem{xia2018dota}
\begin{bchapter}
\bauthor{\bsnm{Xia}, \binits{G.-S.}},
\bauthor{\bsnm{Bai}, \binits{X.}},
\bauthor{\bsnm{Ding}, \binits{J.}},
\bauthor{\bsnm{Zhu}, \binits{Z.}},
\bauthor{\bsnm{Belongie}, \binits{S.}},
\bauthor{\bsnm{Luo}, \binits{J.}},
\bauthor{\bsnm{Datcu}, \binits{M.}},
\bauthor{\bsnm{Pelillo}, \binits{M.}},
\bauthor{\bsnm{Zhang}, \binits{L.}}:
\bctitle{Dota: A large-scale dataset for object detection in aerial images}.
In: \bbtitle{Proceedings of the IEEE Conference on Computer Vision and Pattern
  Recognition},
pp. \bfpage{3974}--\blpage{3983}
(\byear{2018})
\end{bchapter}
\endbibitem

\bibitem{kingma2014adam}
\begin{botherref}
\oauthor{\bsnm{Kingma}, \binits{D.P.}},
\oauthor{\bsnm{Ba}, \binits{J.}}:
Adam: A method for stochastic optimization.
arXiv preprint arXiv:1412.6980
(2014)
\end{botherref}
\endbibitem

\bibitem{everingham2010pascal}
\begin{barticle}
\bauthor{\bsnm{Everingham}, \binits{M.}},
\bauthor{\bsnm{Van~Gool}, \binits{L.}},
\bauthor{\bsnm{Williams}, \binits{C.K.}},
\bauthor{\bsnm{Winn}, \binits{J.}},
\bauthor{\bsnm{Zisserman}, \binits{A.}}:
\batitle{The pascal visual object classes (voc) challenge}.
\bjtitle{International journal of computer vision}
\bvolume{88}(\bissue{2}),
\bfpage{303}--\blpage{338}
(\byear{2010})
\end{barticle}
\endbibitem

\bibitem{IJCAT2012}
\begin{barticle}
\bauthor{\bsnm{Wu}, \binits{J.}},
\bauthor{\bsnm{Cai}, \binits{Z.-h.}},
\bauthor{\bsnm{Ao}, \binits{S.}}:
\batitle{Hybrid dynamic k-nearest-neighbour and distance and attribute weighted
  method for classification}.
\bjtitle{Int. J. Comput. Appl. Technol.}
\bvolume{43}(\bissue{4}),
\bfpage{378}--\blpage{384}
(\byear{2012})
\end{barticle}
\endbibitem

\bibitem{6729567}
\begin{bchapter}
\bauthor{\bsnm{Wu}, \binits{J.}},
\bauthor{\bsnm{Zhu}, \binits{X.}},
\bauthor{\bsnm{Zhang}, \binits{C.}},
\bauthor{\bsnm{Cai}, \binits{Z.}}:
\bctitle{Multi-instance multi-graph dual embedding learning}.
In: \bbtitle{2013 IEEE 13th International Conference on Data Mining},
pp. \bfpage{827}--\blpage{836}
(\byear{2013})
\end{bchapter}
\endbibitem

\bibitem{yang2018automatic}
\begin{barticle}
\bauthor{\bsnm{Yang}, \binits{X.}},
\bauthor{\bsnm{Sun}, \binits{H.}},
\bauthor{\bsnm{Fu}, \binits{K.}},
\bauthor{\bsnm{Yang}, \binits{J.}},
\bauthor{\bsnm{Sun}, \binits{X.}},
\bauthor{\bsnm{Yan}, \binits{M.}},
\bauthor{\bsnm{Guo}, \binits{Z.}}:
\batitle{Automatic ship detection in remote sensing images from google earth of
  complex scenes based on multiscale rotation dense feature pyramid networks}.
\bjtitle{Remote Sensing}
\bvolume{10}(\bissue{1}),
\bfpage{132}
(\byear{2018})
\end{barticle}
\endbibitem

\bibitem{wei2020x}
\begin{barticle}
\bauthor{\bsnm{Wei}, \binits{H.}},
\bauthor{\bsnm{Zhang}, \binits{Y.}},
\bauthor{\bsnm{Wang}, \binits{B.}},
\bauthor{\bsnm{Yang}, \binits{Y.}},
\bauthor{\bsnm{Li}, \binits{H.}},
\bauthor{\bsnm{Wang}, \binits{H.}}:
\batitle{X-linenet: Detecting aircraft in remote sensing images by a pair of
  intersecting line segments}.
\bjtitle{IEEE Transactions on Geoscience and Remote Sensing}
\bvolume{59}(\bissue{2}),
\bfpage{1645}--\blpage{1659}
(\byear{2020})
\end{barticle}
\endbibitem

\bibitem{inproceedings}
\begin{bchapter}
\bauthor{\bsnm{Liu}, \binits{Z.}},
\bauthor{\bsnm{Yuan}, \binits{L.}},
\bauthor{\bsnm{Weng}, \binits{L.}},
\bauthor{\bsnm{Yang}, \binits{Y.}}:
\bctitle{A high resolution optical satellite image dataset for ship recognition
  and some new baselines},
pp. \bfpage{324}--\blpage{331}
(\byear{2017}).
\doiurl{10.5220/0006120603240331}
\end{bchapter}
\endbibitem

\bibitem{xiao2020axis}
\begin{barticle}
\bauthor{\bsnm{Xiao}, \binits{Z.}},
\bauthor{\bsnm{Qian}, \binits{L.}},
\bauthor{\bsnm{Shao}, \binits{W.}},
\bauthor{\bsnm{Tan}, \binits{X.}},
\bauthor{\bsnm{Wang}, \binits{K.}}:
\batitle{Axis learning for orientated objects detection in aerial images}.
\bjtitle{Remote Sensing}
\bvolume{12}(\bissue{6}),
\bfpage{908}
(\byear{2020})
\end{barticle}
\endbibitem

\bibitem{feng2020toso}
\begin{bchapter}
\bauthor{\bsnm{Feng}, \binits{P.}},
\bauthor{\bsnm{Lin}, \binits{Y.}},
\bauthor{\bsnm{Guan}, \binits{J.}},
\bauthor{\bsnm{He}, \binits{G.}},
\bauthor{\bsnm{Shi}, \binits{H.}},
\bauthor{\bsnm{Chambers}, \binits{J.}}:
\bctitle{Toso: Student’st distribution aided one-stage orientation target
  detection in remote sensing images}.
In: \bbtitle{ICASSP 2020-2020 IEEE International Conference on Acoustics,
  Speech and Signal Processing (ICASSP)},
pp. \bfpage{4057}--\blpage{4061}
(\byear{2020}).
\bcomment{IEEE}
\end{bchapter}
\endbibitem

\bibitem{2016Stacked}
\begin{bchapter}
\bauthor{\bsnm{Newell}, \binits{A.}},
\bauthor{\bsnm{Yang}, \binits{K.}},
\bauthor{\bsnm{Jia}, \binits{D.}}:
\bctitle{Stacked hourglass networks for human pose estimation}.
In: \bbtitle{European Conference on Computer Vision}
(\byear{2016})
\end{bchapter}
\endbibitem

\bibitem{2015ICDAR2015}
\begin{bchapter}
\bauthor{\bsnm{Burie}, \binits{J.}},
\bauthor{\bsnm{Chazalon}, \binits{J.}},
\bauthor{\bsnm{Coustaty}, \binits{M.}},
\bauthor{\bsnm{Eskenazi}, \binits{S.}},
\bauthor{\bsnm{Rusinol}, \binits{M.}}:
\bctitle{Icdar2015 competition on smartphone document capture and ocr
  (smartdoc)}.
In: \bbtitle{2015 13th International Conference on Document Analysis and
  Recognition (ICDAR)}
(\byear{2015})
\end{bchapter}
\endbibitem

\bibitem{karatzas2013icdar}
\begin{bchapter}
\bauthor{\bsnm{Karatzas}, \binits{D.}},
\bauthor{\bsnm{Shafait}, \binits{F.}},
\bauthor{\bsnm{Uchida}, \binits{S.}},
\bauthor{\bsnm{Iwamura}, \binits{M.}},
\bauthor{\bparticle{i} \bsnm{Bigorda}, \binits{L.G.}},
\bauthor{\bsnm{Mestre}, \binits{S.R.}},
\bauthor{\bsnm{Mas}, \binits{J.}},
\bauthor{\bsnm{Mota}, \binits{D.F.}},
\bauthor{\bsnm{Almazan}, \binits{J.A.}},
\bauthor{\bsnm{De~Las~Heras}, \binits{L.P.}}:
\bctitle{Icdar 2013 robust reading competition}.
In: \bbtitle{2013 12th International Conference on Document Analysis and
  Recognition},
pp. \bfpage{1484}--\blpage{1493}
(\byear{2013}).
\bcomment{IEEE}
\end{bchapter}
\endbibitem

\end{thebibliography}


\end{document}